\documentclass[a4paper, 10pt, conference]{ieeeconf}      
\usepackage{graphicx}
\usepackage{framed,multirow}

\usepackage{amssymb}
\usepackage{latexsym}

\usepackage{amsmath}
\usepackage{caption}
\usepackage{url}
\usepackage{xcolor}

\IEEEoverridecommandlockouts                              
\title{\LARGE \bf
Skeleton-Based Hand Gesture Recognition by Learning SPD Matrices with Neural Networks
}

\author{\parbox{16cm}{\centering
   {\large Xuan Son Nguyen$^1$, Luc Brun$^1$, Olivier Lezoray$^2$ and S\'ebastien Bougleux$^2$}\\
   {\normalsize
   $^1$ Normandie Univ, ENSICAEN, UNICAEN, CNRS, GREYC, 14000 Caen, France\\
   $^2$ Normandie Univ, UNICAEN, ENSICAEN, CNRS, GREYC, 14000 Caen, France}}
   \thanks{This material is based upon work supported by the European Union and the Region Normandie under the project IGIL.}
}

\begin{document}

\thispagestyle{empty}
\pagestyle{empty}

\maketitle

\begin{abstract}

In this paper, we propose a new hand gesture recognition method based on skeletal data by learning SPD matrices with neural networks.
We model the hand skeleton as a graph and introduce a neural network for SPD matrix learning, taking as input the 3D coordinates
of hand joints. The proposed network is based on two newly designed layers that transform a set of SPD matrices into a SPD matrix.
For gesture recognition, we train a linear SVM classifier using features extracted from our network. 
Experimental results on a challenging dataset (Dynamic Hand Gesture
dataset from the SHREC 2017 3D Shape Retrieval Contest) show that the proposed method outperforms state-of-the-art methods.

\end{abstract}

\section{Introduction}
\label{sec:intro}

Hand gesture recognition is one of the main tasks in video understanding. This task has received increasing attention from the computer vision community,
as it has many potential applications in human-robot interaction, sign language interpretation, and smart home control interface. 
Existing vision-based approaches for hand gesture recognition use generally three data modalities, i.e. RGB and depth (RGB-D) images, and skeletal data.  
Most of these approaches are mainly based on RGB-D images since they can be directly acquired from vision sensors,
while skeletal data must be estimated by a tracking algorithm and is thus more difficult to obtain and less reliable than RGB-D images. 

The introduction of depth cameras like Intel Realsense and Microsoft Kinect enables effective hand gesture recognition using skeletal data~\cite{SmedtCVPRW16,SmedtEuro17}. 
Compared to RGB-D images, skeletal data offers several advantages. First, view-invariant gesture recognition can be achieved 
using skeletal data. Second, features extracted from skeletal data are generally low-dimensional and fast to compute compared to those
extracted from RGB-D images.
Figure~\ref{fig:hand_graph}(a) shows the hand joint positions estimated by an Intel Realsense camera. Motivated by the recent success of graph convolutional networks for action recognition~\cite{LiGraphConvAAAI18,YanAAAI18}, we propose to model the hand skeleton as a graph for hand gesture recognition. 
Figure~\ref{fig:hand_graph}(b) shows the graph at one frame corresponding to the hand skeleton in Fig.~\ref{fig:hand_graph}(a). 
To obtain gesture representations, we rely on first- and second-order statistics. These statistics are combined using
Gaussian distributions~\cite{Li17,G2DeNet17}, forming Symmetric Positive Definite (SPD) matrices. Since SPD matrices are known to lie on a Riemannian manifold, we develop a neural network for SPD matrix learning~\cite{DongSPD17,HuangGool17,Ionescu2015,LiIsSecond17,G2DeNet17}. To this end, we propose two new types of layers that transform a set of SPD matrices into a SPD matrix.
Despite the simplicity of our network, it outperforms state-of-the-art methods on the challenging Dynamic Hand Gesture (DHG) dataset~\cite{SmedtEuro17} from the SHREC 2017 3D Shape Retrieval Contest.
\section{Related Works}
\label{sec:related_work}


\subsection{Skeleton-Based Hand Gesture and Action Recognition}
\label{subsec:ske_approaches}

Ioenescu et al.~\cite{IonescuHand2005} extracted the hand region skeleton and a dynamic signature of the gesture was then computed as the superposition of all the skeletons of the hand region.
Wang and Chan~\cite{WangCip14} segmented the hand region and used a superpixel representation to capture the hand shape and texture information. 
De Smedt et al.~\cite{SmedtCVPRW16} proposed two kinds of features based on hand shapes and movements. The shape-based features were computed using Fisher vector, 
and the movement-based features were computed from histogram of hand directions and histogram of wrist rotations. 
Nunez et al.~\cite{Nez2018} trained a combination of a Convolutional Neural Network (CNN)~\cite{Krizhevsky12ImageNet} 
and a Long Short-Term Memory (LSTM)~\cite{LSTMHochreiter1997} using two stages. In the first stage, 
the CNN was connected to a fully-connected multi-layer perceptron of two hidden layers. In the second stage, the output of the pre-trained CNN was connected to the LSTM. Devineau et al.~\cite{DevineauFG18} separated a gesture sequence into 1D sequences and fed each of them to a multi-channel CNN consisting of three parallel branches. The first branch was a residual network, while the two other branches were designed for feature extraction. 

For skeleton-based action recognition, two categories of features have also been investigated, e.g. hand-crafted features and deep-learned features. 
In hand-crafted feature-based approaches, the 3D joint positions of the human body were used to compute skeletal quad~\cite{SkeQuad14}, 
points in a Lie group~\cite{LieGroup14}, Actionlet Ensemble~\cite{ActionletEns12}, EigenJoints~\cite{EigenJoints}. 
In deep learning based-approaches, various architectures based on CNN, Recurrent Neural Network~\cite{NN15}, 
LSTM~\cite{LiuCVPR17LSTM,Shahroudy16NTU} were introduced to learn action representations.
Very recently, geometric deep learning which extends classical operations like convolution to non-Euclidean domains, e.g. manifolds and graphs, 
has also been applied to skeleton-based action recognition.
Manifold learning approaches operated on different manifolds, 
e.g. SPD manifolds~\cite{HuangGool17}, Lie groups~\cite{Huang17DLLieGroup}, and Grassmann manifolds~\cite{HuangAAAI18}.
Graph learning approaches~\cite{ShiArXiv18,YanAAAI18} were based on graph convolution filters and Deep Residual Networks~\cite{HeResNet16} 
which have shown very promising results on large scale datasets.

\subsection{Deep Learning on Graphs}
\label{subsec:deep_learning_graph}

Deep learning on graphs, as mentioned previously, has received increasing attention in recent years. 
While various approaches have been proposed and proven effective in different 
tasks~\cite{DefferrardCNN2016,SimonovskyEdgeConv17,VermaFeaStNet18}, very few approaches have been developed so far for action and hand gesture recognition. 
Yan et al.~\cite{YanAAAI18} introduced a Spatial Temporal Graph CNN operating on a graph constructed from body joints and physical 
connections between them. Their network was designed and trained based on the mechanism of Deep Residual Networks~\cite{HeResNet16}. 
Shi et al.~\cite{ShiArXiv18} improved~\cite{YanAAAI18} by automatically learning the graph structure at each layer, 
while Zhang et al.~\cite{ZhangGraphEdgeConv18} improved~\cite{YanAAAI18} by defining an equivalent convolution operator on graph edges. 
Similarly to~\cite{YanAAAI18}, Li et al.~\cite{LiGraphConvAAAI18} constructed a graph to represent the human body skeleton. 
A polynomial approximation of the Graph Fourier Transform was then used to design multi-scale convolutional filters,
which were then combined with autoregressive moving average to build a spatio-temporal graphical convolutional model.

\subsection{Deep Learning of SPD Matrices}
\label{subsec:deep_learning_spd} 

Ionescu et al.~\cite{Ionescu2015} introduced the theory and practical techniques for matrix backpropagation
in the deep learning framework. They also proposed DeepO2P, a network 
for region classification, and an algorithm for backpropagation of logarithm of SPD matrices. 
Based on~\cite{Ionescu2015}, several methods for SPD matrix learning 
have been proposed.  
Dong et al.~\cite{DongSPD17} introduced a 2D fully connected layer and a symmetrically clean layer 
for dimensionality reduction and non-linear transformation of SPD matrices.
Huang and Gool~\cite{HuangGool17} mapped a high dimensional SPD matrix into a lower dimensional one via Bimap, a layer 
having a similar form as the 2D fully connected layer proposed in~\cite{DongSPD17}. 
Kernel matrices were also used to design non-linear activation functions~\cite{EnginKspd17}.
Li et al.~\cite{LiIsSecond17} and Wang et al.~\cite{G2DeNet17} combined parametric probability distributions modeling 
into CNNs, which requires backpropagation of a square-rooted SPD matrix.
\begin{figure}[t]
{
\def\a{$\mathbf{w}_1$}
\def\c{$\mathbf{w}_2$}
\def\e{$\mathbf{w}_3$}
\centerline{\scalebox{0.45}{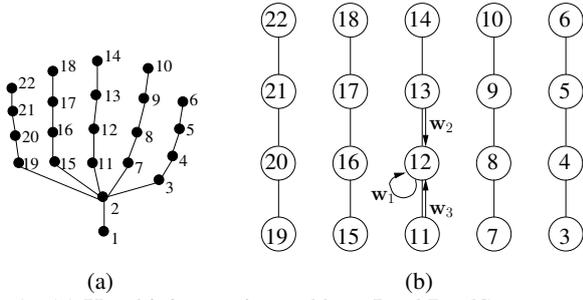}}
\caption{\label{fig:hand_graph} (a) Hand joints estimated by a Intel RealSense camera  
(b) graph of hand skeleton and the weights associated with the neighbors of node 12 in the convolutional layer.} 
}
\end{figure}

\section{Proposed Network}
\label{sec:proposed_method_deep}

\subsection{Overview}\label{subsec:overview}

\begin{figure*}[t]
{
\centerline{\scalebox{0.11}{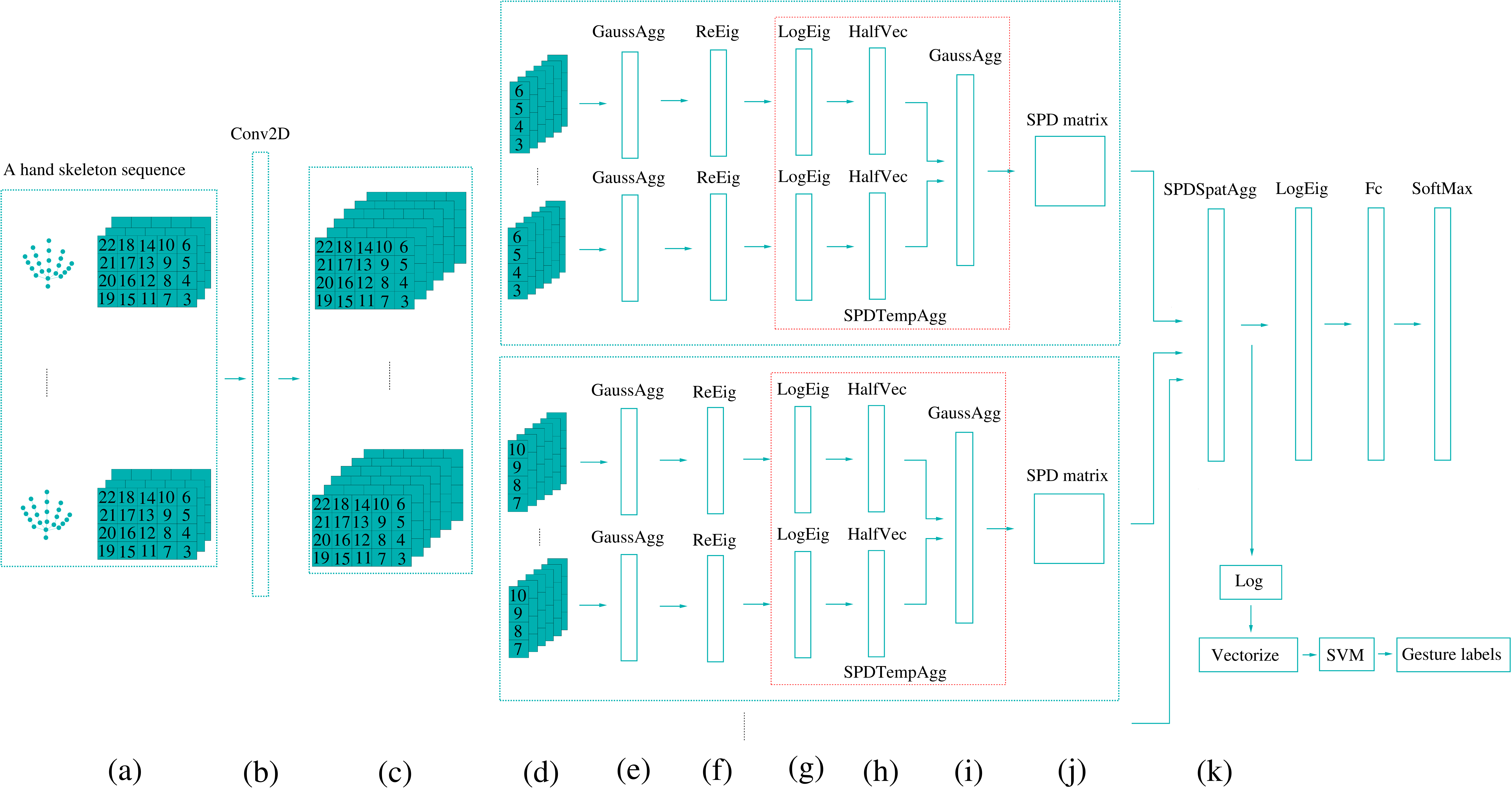}}
\caption{\label{fig:spd_network} (Best viewed in color) The proposed network. 
({\bf a}) A hand skeleton sequence. 
({\bf b}) The convolutional layer for the graph of hand joints. 
({\bf c}) The output of this layer. 
({\bf d}) Construction of different sub-sequences, each of them is associated with a specific finger. Two sub-sequences associated with the thumb and the index are shown here. Each sub-sequence is processed by a branch of the network (from (\textbf{e}) to (\textbf{j})). (\textbf{k}) The SPD matrices at step (\textbf{j}) are transformed into a SPD matrix.
} 
}
\end{figure*}

Given a hand skeleton sequence, we construct a graph for each frame as illustrated in Fig.~\ref{fig:hand_graph}(b). The features at a node provide a 3-dimensional vector formed from the x, y, and z coordinates of the corresponding hand joint.
We design a simple 2D convolutional layer which is applied to the input data to take into account the correlation between neighboring nodes. A sub-sequence of a specific finger is extracted from the sequence and processed by a branch of the network.
Two layers referred to as GaussAgg and SPDTempAgg are then used to learn joint features for each branch. 
The outputs of all branches of the network are combined using a layer referred to as SPDSpatAgg, resulting in a SPD matrix. Since the obtained SPD matrix lies on a Riemannian manifold, it is mapped to Euclidean spaces for classification. The proposed network architecture is illustrated in Fig.~\ref{fig:spd_network}.

In the next two sections, we denote by $n^k_L$ the number of input SPD matrices of the $k^{th}$ layer, by $d_{k-1}$ and $d_k$ the dimensions of input and output SPD matrices  of the $k^{th}$ layer, respectively, by $\mathbf{X}^{i}_{k-1}$, $i=1,\ldots,n_L^k$ the input SPD matrices of the $k^{th}$ layer, and by $\mathbf{X}_k$ the output SPD matrix of the $k^{th}$ layer.

\subsection{SPD Spatial Aggregation} \label{subsec:concat_bimap_layer}
The SPDSpatAgg layer (step ({\bf k}) of Fig.~\ref{fig:spd_network}) computes $\mathbf{X}_k$ by a mapping $f_{sa}$ as:
\begin{align}\label{eq:sumbimap}
\begin{split}
\mathbf{X}_k & = f_{sa}^{(k)} ((\mathbf{X}^{1}_{k-1},\ldots,\mathbf{X}^{n_L^k}_{k-1});\mathbf{W}_k^{1},\ldots,\mathbf{W}_k^{n_L^k}) \\ & = \sum_{i=1}^{n_L^k} \mathbf{W}_k^{i} \mathbf{X}^{i}_{k-1} (\mathbf{W}_k^{i})^T,
\end{split}
\end{align}
where $\mathbf{W}_k^{i} \in \mathbb{R}^{d_k \times d_{k-1}}$ are the transformation matrices.

To guarantee that $\mathbf{X}_k$ is SPD, one can use the same assumption as~\cite{HuangGool17}, i.e. 
each transformation matrix $\mathbf{W}_k^{i}$ is a full row rank matrix. In this case, each term in the right-hand side of Eq.~(\ref{eq:sumbimap}) is SPD, so their sum is also SPD. Then optimal solutions of the transformation matrices are achieved by 
additionally assuming that each transformation matrix $\mathbf{W}_k^{i}$ resides on a compact 
Stiefel manifold $St(d_k,d_{k-1})$\footnote{A compact Stiefel manifold $St(d_k,d_{k-1})$ is the set of $d_k$-dimensional orthonormal matrices of $\mathbb{R}^{d_{k-1}}$.}. 
Note that this assumption imposes a constraint on the dimension of $\mathbf{X}_k$, i.e. $d_k \le d_{k-1}$. The backpropagation procedure of this layer is based on the optimization framework of~\cite{HuangGool17}.

\subsection{SPD Temporal Aggregation}
\label{subsec:spdtemppooling_layer}


The SPDTempAgg layer is decomposed into three sub-layers as illustrated at steps ({\bf g}), ({\bf h}) and ({\bf i})
of Fig.~\ref{fig:spd_network}. 
The first sub-layer computes the matrix logarithm $\mathbf{Y}^{i}_{k-1}$ from $\mathbf{X}^{i}_{k-1}$, for $i=1,\ldots,n_L^k$ as:
\begin{equation}
\mathbf{Y}^{i}_{k-1} = \log (\mathbf{X}^{i}_{k-1}) = \mathbf{U} \log(\mathbf{V}) \mathbf{U}^T,
\end{equation}
where $\mathbf{U} \mathbf{V} \mathbf{U}^T$ is the eigen-decomposition of $\mathbf{X}^{i}_{k-1}$. This layer can be implemented similarly to the LogEig layer~\cite{HuangGool17}.

The second sub-layer, referred to as HalfVec layer, vectorizes the upper triangle of $\mathbf{Y}^{i}_{k-1}$ for $i=1,\ldots,n_L^k$ as:
\begin{align}
\begin{split}
\mathbf{z}^{i} = &[\mathbf{Y}^{i}_{k-1}(1,1), \sqrt{2}\mathbf{Y}^{i}_{k-1}(1,2), \ldots, \sqrt{2}\mathbf{Y}^{i}_{k-1}(1,d_{k-1}), \\ & \mathbf{Y}^{i}_{k-1}(2,2), \sqrt{2}\mathbf{Y}^{i}_{k-1}(2,3),\ldots, \mathbf{Y}^{i}_{k-1}(d_{k-1},d_{k-1})]^T
\end{split}
\end{align}
where $\mathbf{Y}^{i}_{k-1}(u,v)$ denotes the element at the $u^{th}$ row and $v^{th}$ column of $\mathbf{Y}^{i}_{k-1}$.

The third sub-layer, referred to as GaussAgg layer, aggregates the $\mathbf{z}^{i}$, for $i=1,\ldots,n_L^k$, based on their Gaussian distributions~\cite{G2DeNet17} to obtain the output of SPDTempAgg:

\begin{equation}\label{eq:matrix_partition}
\mathbf{X}_k\,{=}\,f_{ta}^{(k)} ((\mathbf{X}^{1}_{k-1},\ldots,\mathbf{X}^{n_L^k}_{k-1}))=\begin{bmatrix} \pmb{\Sigma} + \pmb{\mu} \pmb{\mu}^T & \pmb{\mu} \\ \pmb{\mu}^T & 1 \end{bmatrix}
\end{equation}
where $\pmb{\mu} = \frac{1}{n_L^k} \sum_{i=1}^{n_L^k} \mathbf{z}^{i}$ and 
$\pmb{\Sigma} = \frac{1}{n_L^k} \sum_{i=1}^{n_L^k} (\mathbf{z}^{i} - \pmb{\mu})(\mathbf{z}^{i} - \pmb{\mu})^T$.

The backpropagation procedure of this layer is based on the framework of~\cite{Ionescu2015}.
\subsection{The Network in Detail}\label{subsec:hgd_spd} 
Let $n_J,n_F$ be the number of hand joints and the length of the gesture sequence, respectively. 
The input of our network (Fig.~\ref{fig:spd_network}) is a sequence of hand poses denoted by $\mathbf{p}^t_{0,i} \in \mathbb{R}^3,i=1,\ldots,n_J,t=1,\ldots,n_F$,
where $\mathbf{p}^t_{0,i}$ corresponds to the 3D coordinates of joint $i$ at frame $t$.  
Let $d_1$ be the number of channels of output features in the first layer. Denote by $\mathbf{p}^t_{1,i} \in \mathbb{R}^{d_1},i=3,\ldots,n_J,t=1,\ldots,n_F$ the output of the first layer,
by $p_{1,i,c}^t$ the value of the $c^{th}$ channel of $\mathbf{p}_{1,i}^t$, $c=1,\ldots,d_1$, 
i.e. $\mathbf{p}_{1,i}^t = [p_{1,i,1}^t,\ldots,p_{1,i,d_1}^t]^T$. At the first layer, the value for the $c^{th}$ channel
of the feature vector on node $i$ is given by:
\begin{equation}\label{eq:graph_convolution}
p_{1,i,c}^t = \sum_{j \in \mathcal{N}_i} \mathbf{w}_{l(j,i),c}^T\mathbf{p}_{0,j}^t. 
\end{equation}
where $\mathcal{N}_i$ is the set of neighbors of node $i$, $\mathbf{w}_{l(j,i),c}\in\mathbb{R}^3$ is the filter weight vector for the $c^{th}$ channel, and $l(j,i)$ is defined as:
\begin{equation}\label{eq:vertex_weight_label}
  l(j,i) = \left\lbrace\begin{array}{ll} 1 & \text{if $j-i=0$} \\ 2 & \text{if $j-i=1$} \\ 3 & \text{if $j-i=-1$} \end{array}\right.
\end{equation}
  
At step (d) of Fig.~\ref{fig:spd_network}, we partition the set of hand joints into 5 subsets of joints, each of them contains all joints of a finger. For example, the 5 subsets of joints corresponding to the hand skeleton illustrated in Fig.~\ref{fig:hand_graph} are given by: 
$J_1=\{3,4,5,6\}$, $J_2=\{7,8,9,10\}$, $J_3=\{11,12,13,14\}$, $J_4=\{15,16,17,18\}$, $J_5=\{19,20,21,22\}$. 
Then 5 sequences associated with those 5 subsets are constructed from the output of the first layer. 
For each $J_s$, with $s = 1,\ldots,5$, the GaussAgg layer produces SPD matrices given by:
\begin{equation}\label{eq:spd_elementary}
\mathbf{X}^{s,t}_2 = \begin{bmatrix}  \pmb{\Sigma}^{s,t} + \pmb{\mu}^{s,t}(\pmb{\mu}^{s,t})^T & \pmb{\mu}^{s,t} \\ (\pmb{\mu}^{s,t})^T & 1 \end{bmatrix}
\end{equation}
where $\pmb{\mu}^{s,t}=\frac{1}{4}\sum_{i \in J_s}\mathbf{p}_{1,i}^t$, and  
$\pmb{\Sigma}^{s,t}=\frac{1}{3}\sum_{i \in J_s} (\mathbf{p}_{1,i}^t - \pmb{\mu}^{s,t})(\mathbf{p}_{1,i}^t-\pmb{\mu}^{s,t})^T$.

The next step uses the ReEig layer~\cite{HuangGool17} to perform non-linear transformations of SPD matrices given by:
\begin{equation}\label{eq:reeig}
\mathbf{X}^{s,t}_3 = f_r^{(3)} (\mathbf{X}^{s,t}_2) = \mathbf{U} \max(\epsilon \mathbf{I}, \mathbf{V}) \mathbf{U}^T,
\end{equation}
where $\mathbf{X}^{s,t}_3$, for $s=1,\ldots,5$, $t=1,\ldots,n_F$ are the output SPD matrices, $\mathbf{U} \mathbf{V} \mathbf{U}^T$
is the eigen-decomposition of $\mathbf{X}^{s,t}_2$, 
$\epsilon$ is a rectification threshold, $\mathbf{I}$ is an identity matrix, and $\max (\epsilon \mathbf{I}, \mathbf{V})$
is a diagonal matrix defined as:
\begin{equation}\label{eq:a_definition}
(\max (\epsilon \mathbf{I}, \mathbf{V}))(i,i) = \left\lbrace\begin{array}{ll} \mathbf{V}(i,i) & \text{if } \mathbf{V}(i,i) > \epsilon \\ \epsilon & \text{if } \mathbf{V}(i,i) \le \epsilon. \end{array}\right.
\end{equation}

Next, a multi-level representation is constructed for the sequence associated with $J_s$ for $s=1,\ldots,5$. The sequence is evenly subdivided into $i$ sub-sequences at the $i^{th}$ level, $i=1,\ldots,n_T$, with $n_T$ the total number of levels. 
This creates $n_Q=n_T(n_T+1)/2$ sub-sequences for each $J_s$.
Then the SPDTempAgg layer (Sec.~\ref{subsec:spdtemppooling_layer}) is applied for each sub-sequence to generate SPD matrices $\mathbf{X}^{s,q}_4$, using Eq.~(\ref{eq:matrix_partition}) for $s=1,\ldots,5$ and $q=1,\ldots,n_Q$:
\begin{equation}
\mathbf{X}^{s,q}_4 = f_{ta}^{(4)} ((\mathbf{X}^{s,t^q_b}_3,\ldots,\mathbf{X}^{s,t^q_e}_3)),
\end{equation}
where $t^q_b$ and $t^q_e$ are the beginning and ending frame indices of the $q^{th}$ sub-sequence $(\mathbf{X}^{s,t}_3)$ for ${t=t^q_b,\ldots,t^q_e}$.

Finally, the SPD matrices $\mathbf{X}_4^{s,q}$ are transformed into a SPD matrix by the SPDSpatAgg layer presented in Sec.~\ref{subsec:concat_bimap_layer}. The new matrix is transformed to its matrix logarithm via a LogEig layer before being fed
into a fully connected (FC) layer, followed by a softmax layer.
\subsection{Gesture Representation}
\label{subsec:gesture_recognition}

The final representation of a gesture sequence is generated by extracting features from the output of the SPDSpatAgg layer. 
The obtained matrix is then transformed to its matrix logarithm and finally vectorized.
Let $\mathbf{B} \in \mathbb{R}^d$ be the output matrix of the SPDSpatAgg layer, 
then the final representation of the gesture sequence is given as
$\mathbf{v} = [b_{1,1}, \sqrt{2}b_{1,2}, \sqrt{2}b_{1,3},\ldots, \sqrt{2}b_{1,d+1}, b_{2,2}, \sqrt{2}b_{2,3},\ldots, b_{d,d}]^T$,
where $b_{i,i},i=1,\ldots,d$ are the diagonal entries of $\text{log}(\mathbf{B})$ and $b_{i,j},i<j,i,j=1,\ldots,d$ are the off-diagonal entries 
at the $i^{th}$ row and $j^{th}$ column of $\text{log}(\mathbf{B})$.

\section{Experiments}
\label{sec:exps} 

We conduct experiments using the DHG dataset~\cite{SmedtEuro17}.
The dataset contains 14 gestures performed in two ways: using one finger, and using the whole hand. Each gesture is executed several times by 28 participants. 
The gestures are subdivided into categories 
of fine and coarse : Grab (fine), Tap (coarse), Expand (fine), Pinch (fine), Rotation Clockwise (Rot-CW, fine), 
Rotation Counterclockwise (Rot-CCW, fine), Swipe Right (Swipe-R, coarse), Swipe Left (Swipe-L, coarse), Swipe Up (Swipe-U, coarse), Swipe Down (Swipe-D, coarse), 
Swipe X (Swipe-X, coarse), Swipe V (Swipe-V, coarse), Swipe + (Swipe-+, coarse), Shake (coarse).
The dataset provides the 3D coordinates of 22 hand joints as illustrated in Fig.~\ref{fig:hand_graph}(a). 

\paragraph*{Experimental settings}
The dimension of an output feature vector of the first layer is set to $9$. The dimensions of the transformation matrices of the SPDSpatAgg layer are set to $56 \times 56$. 
The number of levels used for the SPDTempAgg layer is set to $3$. The batch size and the learning rate are set to 30 and 0.01, respectively. The rectification threshold $\epsilon$ used for the ReEig layer is set to 0.0001~[12]. For gesture classification, we use the LIBLINEAR library~[9] with L2-regularized L2-loss (dual) to train the classifier and use the default parameter settings in LIBLINEAR (C is set to 1, the tolerance of termination criterion is set to 0.1 and no bias term is added). 
Following~\cite{SmedtEuro17}, we split the dataset into $1960$ train sequences (70\% of the dataset) and $840$ test sequences (30\% test sequences). 
All sequences of the dataset are normalized to have the same length as the longest sequence, which is $171$. Since a gesture is performed in two ways, the number of gesture classes can be 14 or 28~\cite{SmedtEuro17}. We consider both cases in our experiments. The network trained at epoch 20 is used to extract features (Sec.~\ref{subsec:gesture_recognition}).

\begin{figure}[!t]
  \begin{center}
    \begin{tabular}{c}      
      \includegraphics[width=0.96\linewidth]{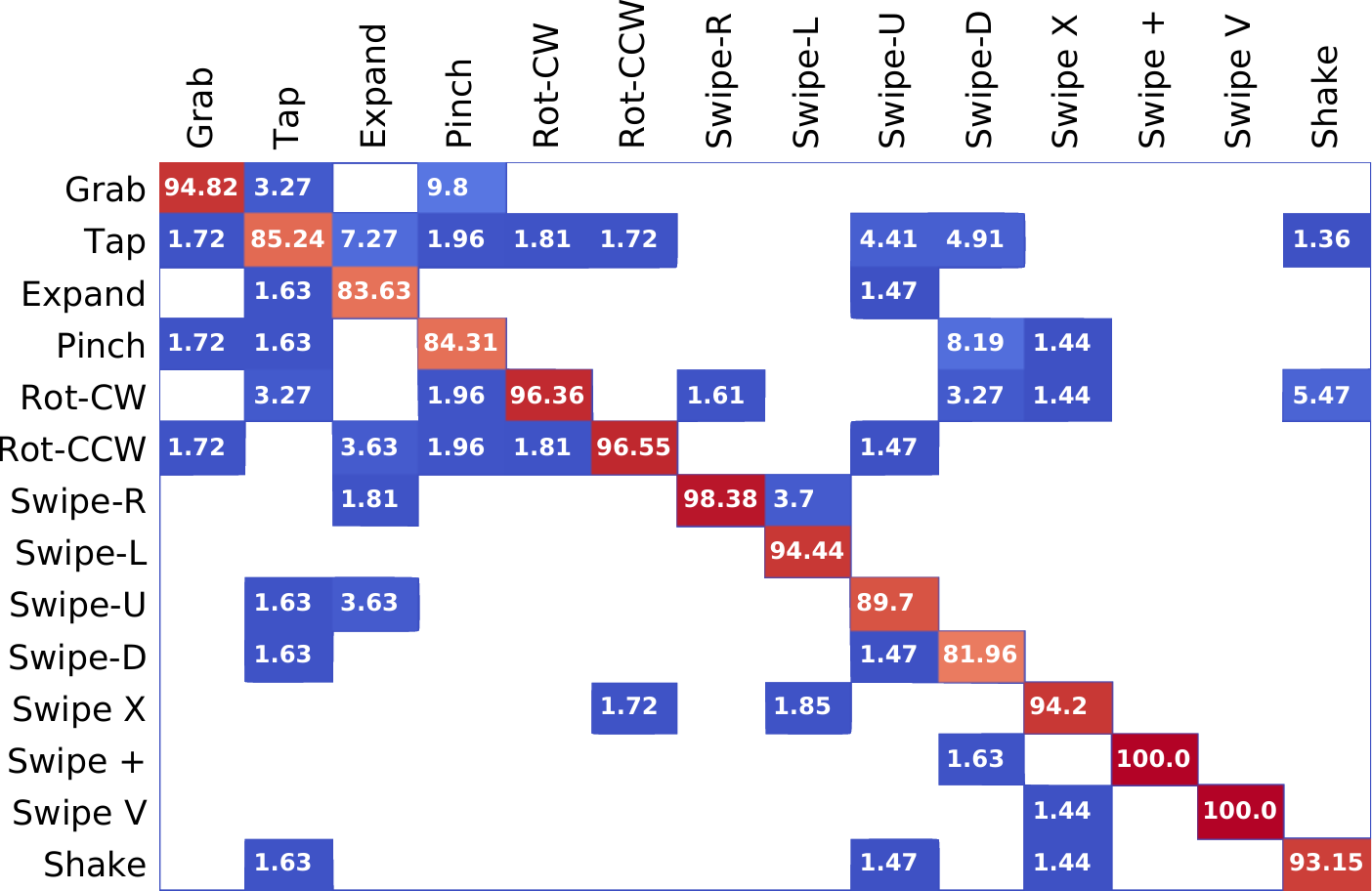} 
    \end{tabular}
  \end{center} 
  \caption{\label{fig:confmat_shrec2017} Confusion matrix for the proposed method on the DHG dataset (best viewed in color).}
\end{figure}

\paragraph*{Results}
Figure~\ref{fig:confmat_shrec2017} shows the confusion matrix of our method on the DHG dataset. 
The recognition errors concentrate on highly similar actions, e.g. Grap to Pinch. 
Our method achieves 100\% accuracy in recognizing actions Swipe-+, Swipe-V, and  more than 90\% accuracy in recognizing actions Grap, Rot-CW, Rot-CCW, Swipe-R, Swipe-L, Swipe-X, Shake. The comparison of our method and state-of-the-art methods on the DHG dataset is given in Tab.~\ref{tab:exp_dhg}.
Our method outperforms the second best method~\cite{DevineauFG18} by 1.1 and 1.96 percent points for experiments with 14 and 28 gestures, respectively. It worth mentioning that our network is much simpler than~\cite{DevineauFG18}. 

\begin{table}[!t]
\begin{center}
  \resizebox{0.75\linewidth}{!}{
  \def\arraystretch{1.2}
  \begin{tabular}{ l  c  c}    
    \hline
     & {\bf 14 classes} & {\bf 28 classes}  \\          
    \hline    
    Huang et al.~\cite{HuangGool17} & 75.24 & 69.64 \\
    Oreifej and Liu~\cite{OreifejHon4d13} & 78.53 & 74.03 \\
    Devanne et al.~\cite{DevanneRie15} & 79.61 & 62.00 \\
    Ohn-Bar and Trivedi~\cite{OhnBar2013} & 83.85 & 76.53 \\
    Chen et al.~\cite{ChenDBLP17} & 84.68 & 80.32 \\
    De Smedt et al.~\cite{SmedtCVPRW16} & 88.24 & 81.90 \\
    Devineau et al.~\cite{DevineauFG18} & 91.28 & 84.35 \\    
    {\bf This paper} & {\bf 92.38} & {\bf 86.31}  \\         
    \hline	
  \end{tabular}
  } 
\end{center}
\caption{\label{tab:exp_dhg} Recognition accuracy comparison on the DHG dataset (best results in bold).}
\end{table}


\section{Conclusions}
\label{sec:conclu}

We have presented a new hand gesture recognition method based on a neural network for SPD matrix learning. 
Our method learns a discriminative SPD matrix encoding the first-order and second-order statistics of local features.
We have provided the experimental evaluation on a benchmark dataset showing the effectiveness of the proposed method. 





\bibliographystyle{ieee} 
\bibliography{references}

\end{document}